\documentclass{article}
\usepackage{ijcai24}

\usepackage{times}
\usepackage{soul}
\usepackage{url}
\usepackage[hidelinks]{hyperref}
\usepackage[utf8]{inputenc}
\usepackage[small]{caption}
\usepackage{graphicx}
\usepackage{amsmath}
\usepackage{amsthm}
\usepackage{booktabs}
\usepackage{algorithm}
\usepackage{algorithmic}
\usepackage[switch]{lineno}
\usepackage{xcolor}
\usepackage{multirow}
\usepackage{amssymb}
\usepackage{rotating}
\usepackage{booktabs}
\usepackage{array}
\usepackage{makecell}
\usepackage{multirow}
\usepackage{pdflscape}
\usepackage{enumitem}

\usepackage{tabularx}

\title{Recent Advances in Predictive Modeling with Electronic Health Records}

\author{
Jiaqi Wang$^1$\and
Junyu Luo$^1$\and
Muchao Ye$^1$\and
Xiaochen Wang$^1$\and
Yuan Zhong$^1$\and
Aofei Chang$^1$\and \\
Guanjie Huang$^1$\and
Ziyi Yin$^1$\and
Cao Xiao$^2$\and
Jimeng Sun$^{3}$\And
Fenglong Ma$^1$\thanks{Corresponding author.}
\affiliations
$^1$Pennsylvania State University\\
$^2$GE Healthcare\\
$^3$University of Illinois Urbana-Champaign\\
\emails
\{jqwang, junyu, muchao, xcwang, yuanzhong, aofei, gzh8, ziyiyin, fenglong\}@psu.edu,\\
cao.xiao@gehealthcare.com,
jimeng@illinois.edu
}

\begin{document}

\maketitle

\begin{abstract}
The development of electronic health records (EHR) systems has enabled the collection of a vast amount of digitized patient data. However, utilizing EHR data for predictive modeling presents several challenges due to its unique characteristics. With the advancements in machine learning techniques, deep learning has demonstrated its superiority in various applications, including healthcare. This survey systematically reviews recent advances in deep learning-based predictive models using EHR data. Specifically, we introduce the background of EHR data and provide a mathematical definition of the predictive modeling task. We then categorize and summarize predictive deep models from multiple perspectives. Furthermore, we present benchmarks and toolkits relevant to predictive modeling in healthcare. Finally, we conclude this survey by discussing open challenges and suggesting promising directions for future research.

   
\end{abstract}

\section{Introduction}

 

Advancements in healthcare technology have sparked a revolution in patient information storage. Many health institutions and providers are now adopting electronic health record (EHR) systems to digitize patient information. This enables healthcare professionals to extract valuable insights from EHR data, thereby supporting various aspects of healthcare, ranging from enhancing clinical decision-making to predicting diseases, enabling personalized medicine, driving quality improvement initiatives, and contributing to public health surveillance and epidemiological research.

This survey focuses on \textbf{predictive modeling} in healthcare, which refers to using machine learning techniques to analyze patients' historical health data along with current observations to support diagnosis or make predictions about future health events. This approach usually leverages large datasets, often derived from EHR systems, to identify patterns, trends, and relationships that can be used to forecast specific health-related events. Predictive modeling has various applications in healthcare, including but not limited to disease risk assessment, hospital discharge potential or readmission risk, treatment outcomes prediction, and health resource allocation. In other words, predictive modeling can enhance clinical decision-making, improve patient care, and contribute to more efficient and cost-effective healthcare delivery.

However, accurate and reliable predictive modeling in healthcare is challenging  due to the unique characteristics of EHR data, summarized as follows:
\begin{itemize}[leftmargin=*]
    \item \textbf{Temporal Dynamics}: The sequentially collected EHR data  exhibits temporal dynamics. Changes in a patient's health status over time may not follow a linear pattern, wihle capturing temporal dependencies requires sophisticated modeling techniques.
    \item \textbf{High Dimensionality}: Medical codes in EHR data (e.g., diagnosis, medication, procedure codes) are high-dimensional. When encoded using multihot encoding,  the codes that are related to the prediction task are often sparse.
    \item \textbf{Multimodalities and Heterogeneity}: Health data comprises of multiple modalities such as clinical notes, laboratory results, imaging reports, and administrative records. The data could also be assembled from different sources. The heterogeneity of the shapes and sources of the data can bring challenges to data integration, which is a foundational task before predictive modeling. 
    \item \textbf{Imbalanced Data}: In healthcare, some patient cohorts could be far smaller than certain majority groups, while certain outcomes or events may be rare, leading to imbalances in the distribution of classes. This can impact the performance of predictive models, making them biased toward the majority class.
    \item \textbf{Clinical Explainability}: Healthcare professionals often require predictive models to be explainable. While effective, many advanced machine learning algorithms may lack explainability, making it challenging for clinicians to understand, trust, and utilize the model results.
\end{itemize}

In this comprehensive survey, we present a systematic overview of recent advancements in deep learning techniques to enhance the performance of predictive models by addressing the aforementioned challenges. The contributions of this paper can be summarized as follows: (1) We introduce a taxonomy based on the techniques employed in existing deep learning-based predictive modeling approaches, offering a detailed analysis of current methods in Section~\ref{sec:3}. (2) Common benchmark datasets and toolkits are summarized to provide a valuable resource for researchers and practitioners in Section~\ref{sec:4}. (3) We discuss open challenges and propose future research directions in the field of predictive modeling in Section~\ref{sec:5}.

\section{Background}\label{sec:2}
\paragraph{EHR Data.}
Let $\mathcal{D} = \{\mathcal{P}_1, \cdots, \mathcal{P}_N\}$ denote the extracted ataset from an EHR system, where $\mathcal{P}_n$ represents the $n$-th patient information. EHR data consists of a sequence of a patient's visit information associated with a time stamp, which can be represented as $\mathcal{P}_n = \{(\mathcal{V}^n_1, t^n_1), \cdots, (\mathcal{V}^n_T, t^n_T)\}$, where $\mathcal{V}_i^n$ and $t_i^n$ denotes the visit information and the time information, respectively, and $T$ is the number of visits. Due to the heterogeneity of EHR data, each visit $\mathcal{V}_i^n$ usually contains $M$ modalities, denoted as $\mathcal{V}_i^n = \{\mathcal{M}_{i,j}^n\}_{j=1}^M$, where each modality $\mathcal{M}_{i,j}^n$ can be either demographical information, diagnosis codes, medication codes, procedure codes, images, time-series monitoring data, or even a clinical note. 

\paragraph{Predictive Modeling.}
The predictive modeling task is formulated as follows: Given the historical EHR data of the $n$-th patient as represented below,
\begin{equation*}\small
\begin{split}
    \mathcal{P}_n &= \{(\mathcal{V}^n_1, t^n_1), \cdots, (\mathcal{V}^n_T, t^n_T)\} \\
    &= \{([\mathcal{M}_{1,1}^n, \cdots, \mathcal{M}_{1,M}^n], t^n_1), \cdots, ([\mathcal{M}_{T,1}^n, \cdots, \mathcal{M}_{T,M}^n], t^n_T)\},
\end{split}
\end{equation*}
the goal is to predict the future health event or outcome $y_n$ after a specified time window (e.g., six months or one year after the last timestamp $T$).
The applications of predictive modeling tasks typically involve binary classification. The cross-entropy (CE) loss is employed to train the models:
\begin{equation*}\small
    \mathcal{L} = \sum_{n=1}^N \text{CE}(f(\mathcal{P}_n; \Theta), y_n),
\end{equation*}
where $f(\cdot)$ represents the model, and $\Theta$ denotes the model parameter set. In the following sections, we will summarize recent advancements in the development of accurate deep learning-based predictive models (i.e., $f(\cdot)$) in healthcare.



\section{Existing Progress}\label{sec:3}
We summarize the recent work on predictive modeling from different perspectives, as shown in Table~\ref{tab:sum}. In the following subsections, we provide details of each type of existing work.

\begin{table*}
    \centering

     \resizebox{\linewidth}{!}{%
    \begin{tabular}{l|c||c|c|c|c|c|c|c||c|c}
    \toprule
     \multirow{2}{*}{\textbf{Model Name}}   &  \multirow{2}{*}{\textbf{Venue}} & \multicolumn{7}{c||}{\textbf{Model Focus}} & \multirow{2}{*}{\textbf{Tasks}} & \multirow{2}{*}{\textbf{Datasets}} \\\cline{3-9}
     & & F1 & F2 & F3 & F4 & F5& F6 & F7 & &\\\hline

     NNR~\cite{mlpResample} & \makecell{JMBE} & & & & &  & & \checkmark & Diabetes & Private\\\hline

     Retain~\cite{choi2016retain}    & NeurIPS & \checkmark & & & \checkmark & & & & Heart Failure & Private\\\hline

     Dipole~\cite{ma2017dipole} & KDD & \checkmark & & & \checkmark & & & & Diagnosis Prediction & Private\\\hline
     T-LSTM~\cite{baytas2017patient}& KDD & \checkmark &  & {} & {} & {} & {} & {} & \makecell{Diabetes Mellitus\\Parkinson’s Progression} & \makecell{EMRBots, PPMI} \\\hline
     GRAM~\cite{choi2017gram} & KDD & \checkmark & & \checkmark & \checkmark & & & & Dignosis Prediction & MIMIC, Private\\\hline

     Timeline~\cite{bai2018interpretable}& KDD & \checkmark & {} & {} & \checkmark & {} & {} & {} &  Admission & SEER \\\hline
    Health-ATM~\cite{ma2018health}&SDM& \checkmark & \checkmark & {} &   & {} & {} & {} & Congestive Heart Failure& Private, EMRbots\\\hline
    RetainEX~\cite{kwon2018retainvis}& TVCG & \checkmark &  & {} & \checkmark & {} & {} & {} & \makecell{Heart Failure\\Cataract} & HIRA \\\hline
    KAME~\cite{ma2018kame}& CIKM & \checkmark & & \checkmark & \checkmark & & & & Dignosis Prediction & MIMIC, Private\\\hline
    PRIME~\cite{ma2018risk}& KDD & \checkmark & & \checkmark &  & & & & \makecell{Heart Failure, COPD\\kidney Disease} & Private\\\hline
    RAIM \cite{xu2018raim} & KDD & \checkmark & & &  & \checkmark & & & \makecell{Decompensation\\ Length of Stay} & MIMIC\\\hline

     DCMN \cite{feng2019dcmn} & ICDM & \checkmark & & &  & \checkmark & & & \makecell{Mortality, Cost} & MIMIC, Private\\\hline
     K-Boosted C5.0~\cite{breastundersmaple} & JHE & & & & & & & \checkmark & Breast Cancer & Private\\\hline

     Stage-Net~\cite{gao2020stagenet}& WWW & \checkmark & {} & {} &   & {} & {} & {} & \makecell{Decompensation\\Mortality}& \makecell{MIMIC, ESRD} \\\hline
     BEHRT~\cite{li2020behrt}& \makecell{Scientific\\Reports} & \checkmark & {} & {} & \checkmark & {} & {} & {} & \makecell{Disease Prediction\\Diesease Embedding} & CPRD \\\hline
     HiTANet~\cite{luo2020hitanet} & KDD & \checkmark & \checkmark & {} & \checkmark & {} & {} & {} & Disease Prediction & Private \\\hline
     Concare~\cite{ma2020concare}& AAAI & \checkmark & \checkmark & {} & \checkmark & {} & {} & {} &  Mortality & MIMIC, Private \\\hline
     F-LSTM \& F-CNN~\cite{tang2020democratizing} & JAMIA & \checkmark & & &  & \checkmark & & & \makecell{Acute Renal Failure\\Shock, Mortality} & MIMIC\\\hline
     MaskEHR~\cite{maskehr} & SDM & \checkmark & & & & & & \checkmark & \makecell{Pastic  Quadriplegia  CerebralPalsy, \\Quadriplegia, Diplegic  Cerebral  Palsy}  & Private\\\hline
     medGAN~\cite{medgan} & MLHC & & & & & & & \checkmark & Heart Failure & \makecell{Private, MIMIC}\\\hline
     LSAN~\cite{ye2020lsan}& CIKM & \checkmark & & &  \checkmark& & & & \makecell{Heart Failure\\ COPD, Kidney Disease} & Private\\\hline

     MedPath~\cite{ye2021medpath} & WWW& \checkmark &  & \checkmark& \checkmark & & & & \makecell{Heart Failure, COPD\\Kidney Disease}& Private\\ \hline
     MedRetriever~\cite{ye2021medretriever} & CIKM & \checkmark & &\checkmark & \checkmark & & & & \makecell{Heart Failure, COPD\\Kidney Disease}&Private \\\hline
     LstmBert~\cite{yang2021leverage} & EMNLP & \checkmark & & &  & \checkmark & & & \makecell{Acute Renal Failure\\ Diagnoses} & MIMIC, Private\\\hline
     MUFASA \cite{xu2021mufasa} & AAAI &\checkmark & & & &  \checkmark & \checkmark &  & Diagnoses & MIMIC\\\hline
     synTEG~\cite{synteg} & JAMIA & \checkmark & & & &  & & \checkmark & \makecell{Type-2 diabetes, Heart Failure,\\ Hypertension, COPD} & Private\\\hline
     EVA~\cite{eva} & MLHC & \checkmark & & & & & & \checkmark & Heart Failure & Private\\\hline
       ConCAD~\cite{huang2021concad} & ECMLPKDD &\checkmark  & & \checkmark&  & &  & &Sleep Apnea &  \makecell{Apnea-ECG, MIT-BIH PSG}\\ \hline

     TContextGGAN~\cite{xu2022time}& TKDE & \checkmark & \checkmark & {} &  & {} & {} & {} &  \makecell{Heart Failure\\Sepsis Risk} & MIMIC \\\hline
     FedCovid~\cite{wang2022towards}& ECMLPKDD&\checkmark&&&&\checkmark&&\checkmark& \makecell{Covid-19 Vaccine Side Effects}& Private\\\hline
     AutoMed \cite{cui2022automed} & BIBM & \checkmark& & & &  & \checkmark &  & \makecell{COPD, Amnesia\\Kidney Disease} & Private\\\hline
     promptEHR~\cite{promptehr} &EMNLP& \checkmark & & & & \checkmark & & \checkmark & Heart Failure & MIMIC\\\hline
     AccSleepNet~\cite{huang2022accsleepnet} & BIBM &\checkmark  & & &  & &  & &Sleep Stage Scoring &  \makecell{Sleep-Accel, Newcastle-Accel}\\\hline
     TrustSleepNet~\cite{huang2022accsleepnet} & BHI &\checkmark  & & &  & \checkmark &  & & Sleep Stage Scoring &  \makecell{SHHS}\\\hline

     KerPrint~\cite{yang2023kerprint}& AAAI& \checkmark & &\checkmark & \checkmark & & & & Disease Prediction & MIMIC, Private\\\hline
     MedHMP~\cite{wang2023hierarchical} & EMNLP & \checkmark & & &  & \checkmark & & & \makecell{Acute Renal Failure, \\Shock, Mortality, Readmission, \\Heart Falire, COPD, Amnesia}& MIMIC, Private\\\hline
     SASMOTE~\cite{smote} & \makecell{BioData\\Mining} & & & & &  & & \checkmark & \makecell{Autism Spectrum Disorder, \\Congenital Heart Disease} & Private\\\hline
     TWIN~\cite{twin} & KDD & \checkmark & & & & \checkmark & & \checkmark & \makecell{Breast Cancer, Lung Cancer} & \makecell{Project Data Sphere}\\\hline

     AutoFM~\cite{cui2024automated} & SDM & \checkmark& & & &  \checkmark& \checkmark &  & \makecell{Acute Renal Failure\\Diagnoses} & MIMIC\\\hline
     MedDiffusion~\cite{meddiff} & SDM & \checkmark & & & & & & \checkmark & \makecell{COPD,Heart Failure,\\Amnesia,Kidney} & \makecell{MIMIC, Private}\\ 


     

     \bottomrule
    \end{tabular}
    }
\caption{Summarization of recent work on predictive modeling in healthcare (in publication years). \textbf{F1}: modeling temporality; \textbf{F2}: modeling time irregularity; \textbf{F3}: incorporating extra knowledge; \textbf{F4}: interpretability; \textbf{F5}: modeling multimodal EHR data; \textbf{F6}: using automated machine learning (AutoML) techniques; and \textbf{F7}: addressing class imbalance issues.}
    \label{tab:sum}
    
\end{table*}



\subsection{Basic Deep Learning-Based Predictive Models} 
\paragraph{Recurrent Neural Networks-Based Models.}  
Due to the temporality of EHR data, recurrent neural networks (RNN) are the primary structure choice in deep learning for handling temporality. In general, predictive models of this type regard EHR data as text-like sequence data and utilize RNNs such as Long Short-Term Memory (LSTM) and Gated Recurrent Units (GRUs) to propagate each visit information to obtain a comprehensive representation. An early attempt at this design named Retain~\cite{choi2016retain} is proposed by Choi et al., which augments RNNs with an attention mechanism and uses those learned attention scores for interpretability.
RNN-based predictive models can even act as a doctor and predict the diagnosis and medication for a future visit given the previous input diagnosis and medication codes~\cite{choi2016doctor}. Later, Dipole~\cite{ma2017dipole} is proposed based on the bidirectional RNN, which aggregates the input EHR data and can effectively attend features in the hierarchy of visits for health risk prediction. These successful early attempts show the effectiveness of the design of combining RNNs with attention mechanisms, and other designs in this fashion come after that, including   Timeline~\cite{bai2018interpretable}, Stage-Net~\cite{gao2020stagenet} and Health-ATM~\cite{ma2018health}.

\paragraph{Transformer-Based Models.}
Recent advancements have highlighted the Transformer architecture~\cite{vaswani2017attention}, which leverages the self-attention mechanism to model relationships within sequential data. Pioneering studies using Transformers in healrhcare such as BEHRT~\cite{li2020behrt} and HiTANet~\cite{luo2020hitanet}, among others~\cite{rasmy2021med,li2022hi,yang2023transformehr,huang2022accsleepnet}, exemplify this trend. BEHRT~\cite{li2020behrt} innovatively utilizes a transformer instead of the conventional RNN module to process sequential EHR records, though it lacks a hierarchical approach for handling code-visit level data. In contrast, HiTANet~\cite{luo2020hitanet} melds the Transformer architecture with hierarchical modeling. It includes a local evaluation stage, employing a time-aware Transformer for embedding temporal aspects into visit-level data, and a global synthesis stage, leveraging a time-aware key-query attention mechanism for a comprehensive analysis.
Adopting Transformer-based models in EHR predictive modeling marks a significant paradigm shift from traditional RNNs. These models offer a more sophisticated approach to managing the intricate and sequential nature of EHR data, demonstrating their immense potential in enhancing healthcare predictive analytics.

\subsection{Time-Aware Predictive Modeling} 
A critical aspect of predictive modeling with EHR data is the integration of temporal information. Contrasting with textual data, the sequence of EHR data hinges on time information, and the intervals between recordings often vary. Accurately modeling this time aspect is essential for evaluating the impact of each patient visit.
Addressing this need, T-LSTM~\cite{baytas2017patient} was introduced, incorporating an information decay function that acknowledges potential decay in patient information over time gaps between visits. This approach modifies the gates of LSTM to enhance risk prediction accuracy.
Building on this concept, several models have emerged. RetainEX~\cite{kwon2018retainvis}, an extension of the Retain model, considers information decay and employs traditional attention mechanisms to assign weights to visits and diagnosis codes. TimeLine~\cite{bai2018interpretable} implements self-attention mechanisms to boost performance and integrate the information decay function into patient representation learning.
Further advancing this field, HiTANet~\cite{luo2020hitanet} introduces a non-monotonic time attention mechanism tailored to disease-specific time preferences using a learnable time preference embedding. Concare~\cite{ma2020concare} enhances time modeling through a multi-channel time series embedding, capturing complex temporal relationships. Additionally, T-ContextGGAN~\cite{xu2022time} employs graph neural networks to model intricate time paths on graphs.
The evolution of time-aware predictive models in EHR data has significantly progressed from simple time information integration to complex, disease-specific time preference modeling. These developments underscore the importance of temporal dynamics in predictive healthcare analytics, leading to more accurate and personalized patient care predictions.

\subsection{Knowledge-Enhanced Predictive Modeling} 
A critical avenue for enhancing predictive models in the EHR domain lies in integrating additional external knowledge. Below, we categorize the methods for leveraging external knowledge based on the type of knowledge being utilized.

\paragraph{Structured Knowledge.}
GRAM~\cite{choi2017gram} utilizes a graph attention network to integrate hierarchical medical ontologies into the learning process of code representations. Building upon GRAM, KAME~\cite{ma2018kame} advances this concept by embedding ontology information throughout the entire prediction process, thereby enriching the model's contextual understanding.
Beyond merely serving as an additional feature, external knowledge can also function as a regularization component. For instance, PRIME~\cite{ma2018risk} generates a posterior regularization term from external knowledge, aligning the model's predictions with established medical insights. This strategy simplifies incorporating external knowledge, making it broadly applicable across various EHR risk prediction frameworks. Besides graph attention, ConCAD~\cite{huang2021concad} develops a cross-attention mechanism to combine deep representations with expert knowledge.

MedPath~\cite{ye2021medpath} introduces a novel approach by retrieving personalized external information from knowledge graphs, thereby enhancing the relevance and utility of the external knowledge. It constructs a personalized ontology map based on patient-specific data, which is then encoded via a graph network for application in predictive tasks. Similarly, KerPrint~\cite{yang2023kerprint} employs both a personalized local knowledge graph based on temporal data and a global-level knowledge graph, optimizing the usage of external information. Subsequent methodologies extend the application of this external knowledge to diverse scenarios, such as the cold-start setting~\cite{tan2022metacare++} and treatment recommendation systems~\cite{yao2023ontology}.

\paragraph{Unstructured Knowledge.}
In contrast, some researchers focus on harnessing unstructured data, such as textual information, which presents unique challenges due to its unorganized nature. MedRetriever~\cite{ye2021medretriever} innovatively addresses this by creating a pool of medical text segments from various online sources. It leverages EHR data and the target disease document embeddings to dynamically retrieve relevant text segments, offering insights into the progression of a disease from symptoms to diagnosis.


\subsection{Predictive Modeling with Multimodal Data} 
Compared to conventional methods that rely solely on a single modality, multimodal approaches leverage diverse clinical modalities as input. This enables them to gather comprehensive information, leading to more promising performance outcomes.

\paragraph{Centralized Learning.}
The prior study by Tang et al.~\cite{tang2020democratizing} suggests the fusion of demographics and temporal clinical features through concatenation in the early stages, followed by conventional machine learning models for the ultimate prediction. In contrast, DCMN~\cite{feng2019dcmn} employs a dual memory network to concurrently process two input modalities, namely waveform and clinical sequence. The outputs of both memory networks are then summed to facilitate further predictions.

RAIM~\cite{xu2018raim} integrates waveform and vital signs as inputs into a multi-channel attention mechanism module. The module subsequently aggregates discrete clinical features into the generated embeddings for subsequent prediction tasks. Another approach, proposed by Yang et al.~\cite{yang2021leverage}, focuses on fusing temporal clinical features, time-invariant data, and clinical notes through summation in the output stage. The researchers explore various combinations of modality-specific encoders to achieve competitive prediction performance.

A more recent contribution by Wang et al.~\cite{wang2023hierarchical} leverages an attention mechanism to obtain a weighted summation of multiple clinical modalities, including temporal clinical features, diagnosis, medication, clinical notes, and demographics. This approach aims to generate a meaningful multimodal Electronic Health Record (EHR) representation for both pretraining and task-specific fine-tuning.

\paragraph{Federated Learning.}
To protect EHR data privacy, the multimodal EHR data is usually stored in local servers and not shared with others. Thus, collaborative training of predictive models without sharing EHR data is a practical and challenging task. 
FedCovid~\cite{wang2022towards}, a federated learning framework, is developed to adaptively fuse heterogeneous EHR data at local client sites for predicting COVID-19 vaccine side effects. 
Besides, \cite{sachin2023multimodal} introduces a multimodal contrastive federated learning framework for digital healthcare, employing a geometric multimodal contrastive representation method to enhance the representation of various modalities in a shared space, thus improving inter-modal relationship capture and overall model performance.

\subsection{AutoML-Based Predictive Modeling}
EHR typically encompasses structured and unstructured data with sparse and irregular longitudinal features. In multimodal fusion, the challenge lies in determining how to effectively fuse different modalities, a problem that often relies on manual modeling and intuition. We have encountered difficulties choosing between early, hybrid, and late fusion approaches. Recently, automated machine learning (AutoML)-based approaches have been proposed to search for optimal fusion strategies automatically.

MUFASA~\cite{xu2021mufasa} is the first work that applied the neural architecture search (NAS) technique to medical multimodal fusion. It jointly optimizes multimodal fusion strategies and modality-specific architectures. MUFASA delineates two types of blocks: modality-specific block, which only allows inputs from the same modality, and fusion block, which accepts inputs from both fusion architecture hidden states and modality-specific states. AutoMed~\cite{cui2022automed} employs a general Directed Acyclic Graph (DAG) framework, opting for a cell-based search where each cell consists of a sequentially ordered set of computation nodes. It uses the same search space design for modality encoding cells and fusion cells. Another noteworthy work, AutoFM~\cite{cui2024automated}, also focuses on modality-specific search and multimodal fusion search. This approach designs separate search spaces for static and sequential modalities. Furthermore, it incorporates modality interaction operations within these modality-specific search spaces to facilitate early interactions. During the fusion search phase, a specialized set of simple fusion operations is devised to enable effective feature fusion.

\subsection{Predictive Modeling with Imbalanced Classes} 
Addressing the issue of class imbalance is another critical challenge in healthcare predictive modeling. Such imbalance can significantly degrade the performance of predictive models. To tackle this issue, a variety of strategies have been developed and extensively researched. In this section, we systematically categorize and review recent methodologies, providing a structured overview of the techniques employed to effectively manage class imbalance in healthcare datasets.

\paragraph{Oversampling and Undersampling Techniques.}
Oversampling techniques are crucial for balancing datasets, effectively increasing the size of the minority class. SASMOTE~\cite{smote} that builds on the Synthetic Minority Oversampling Technique (SMOTE) proves effective in enhancing model accuracy for imbalanced gene risk and heart disease datasets. Furthermore, deep learning-based resampling NNR~\cite{mlpResample} shows great success in imbalanced diabetes prediction. Compared to oversampling, undersampling works by selecting informative samples near the boundary of the class. For instance, K-Boosted C5.0~\cite{breastundersmaple} utilizes K-means to segregate the majority and minority classes and select an equal amount of patients per cluster for Breast Cancer prediction.

\paragraph{Generative Techniques.}
Generative models, primarily developed for enhancing entire datasets, can be specifically tailored to augment minority data. MaskEHR~\cite{maskehr}, medGAN~\cite{medgan}, and synTEG~\cite{synteg} employ Generative Adversarial Networks (GANs) for generating synthetic data. EVA~\cite{eva} and TWIN~\cite{twin} use Variational Autoencoders (VAEs) in their approach. MedDiffusion~\cite{meddiff} leverages the capabilities of Denoising Diffusion Probabilistic Models (DDPMs). Additionally, promptEHR~\cite{promptehr} utilizes a language model for similar purposes. These models use sequential learning techniques like RNNs and Transformers to add historical context, which maintains the temporal coherence of generated data. 
By incorporating this augmented minority class data into the training sets, predictive models benefit significantly, exhibiting enhanced performance and more balanced outcomes.


\begin{table*}[t]
    \centering
    \resizebox{\linewidth}{!}{
    \begin{tabular}{c|c|c|p{0.35\linewidth}|c}
    \toprule
       \textbf{Name}   & \textbf{Data Type}  & \textbf{\# of Data} & \textbf{Modalities} & \textbf{Link}\\\hline
       
       \multirow{6}{*}{MIMIC-III}   & \multirow{6}{*}{Real} &  \multirow{6}{*}{38,597 patients} & Demographics, vital signs, medications, laboratory measurements, observations and notes charted by care providers, fluid balance, procedure codes, diagnostic codes, imaging reports, hospital length of stay, survival data & \multirow{6}{*}{\url{https://physionet.org/content/mimiciii/1.4/}}\\\hline
       
       \multirow{6}{*}{MIMIC-IV}   & \multirow{6}{*}{Real} &  \multirow{6}{*}{40,000+ patients} & Demographics, vital signs, medications, laboratory measurements, observations and notes charted by care providers, fluid balance, procedure codes, diagnostic codes, imaging reports, hospital length of stay, survival data & \multirow{6}{*}{\url{https://physionet.org/content/mimiciv/2.2/}}\\\hline
       
       \multirow{3}{*}{MIMIC-CXR}   & \multirow{3}{*}{Real} &  \multirow{3}{*}{\makecell{377,110 images \\ 227,835 reports}} & Electronic health record data, images (chest radiographs), and natural language (free-text reports) & \multirow{3}{*}{\url{https://physionet.org/content/mimic-cxr/2.0.0/}}\\\hline
       
       \multirow{4}{*}{eICU}   & \multirow{4}{*}{Real} &  \multirow{4}{*}{\makecell{200,000+ admissions}} & Vital sign measurements, care plan documentation, severity of illness measures, diagnosis information, treatment information, and more & \multirow{4}{*}{\url{https://physionet.org/content/mimic-cxr/2.0.0/}}\\\hline

        \multirow{2}{*}{PPMI}   & \multirow{2}{*}{Real} &  \multirow{2}{*}{2,230 patients} & Subject characteristics, biospecimen, images, medical history, etc.& \multirow{2}{*}{\url{https://www.ppmi-info.org/}}\\\hline
        
        \multirow{2}{*}{ADNI}   & \multirow{2}{*}{Real} &  \multirow{2}{*}{2,775 patients} & Subject characteristics, genetic data, images, medical history, neuropathology, etc.& \multirow{2}{*}{\url{https://adni.loni.usc.edu/}}\\\hline

        \multirow{1}{*}{Apnea-ECG}   & \multirow{1}{*}{Real} &  \multirow{1}{*}{70 recordings} & Subject characteristics, electrocardiogram& \multirow{1}{*}{\url{https://physionet.org/content/apnea-ecg/1.0.0/}}\\\hline

        \multirow{3}{*}{MIT-BIH PSG}   & \multirow{3}{*}{Real}  & \multirow{3}{*}{18 recordings} & Subject characteristics, electrocardiogram, electroencephalography, electrooculography, electromyography, etc.& \multirow{3}{*}{\url{https://physionet.org/content/slpdb/1.0.0/}}\\\hline

        \multirow{3}{*}{SHHS}  & \multirow{3}{*}{Real}  & \multirow{3}{*}{6,441 patients} & Subject characteristics, electrocardiogram, electroencephalography, electrooculography, electromyography, airflow, etc.& \multirow{3}{*}{\url{https://sleepdata.org/datasets/shhs}}\\\hline

        \multirow{2}{*}{Newcastle-Accel}  & \multirow{2}{*}{Real}  & \multirow{2}{*}{28 patients} & Subject characteristics, acceleration, polysomnography. & \multirow{2}{*}{\url{https://zenodo.org/records/1160410\#\#.YLqiSC1h1qt}}\\\hline
        
        \multirow{1}{*}{Sleep-Accel}  & \multirow{1}{*}{Real}  & \multirow{1}{*}{31 patients} & Acceleration, heart rate, steps. & \multirow{1}{*}{\url{https://physionet.org/content/sleep-accel/1.0.0/}}\\\hline

        \multirow{2}{*}{EMRBOTS}   & \multirow{2}{*}{Synthetic} &  \multirow{2}{*}{100,000 patients} & Patients’ admissions, demographics, socioeconomics, labs, medications, etc.& \multirow{2}{*}{\url{http://www.emrbots.org/}}\\\hline

        \multirow{2}{*}{Project Data Sphere}   & \multirow{2}{*}{Real} &  \multirow{2}{*}{242 studies} & Data provider, sponsor, study phase, linked data, tumor type, access, etc.& \multirow{2}{*}{\url{https://www.projectdatasphere.org}}\\

        \bottomrule
        
    \end{tabular}}
    \vspace{-0.1in}
    \caption{Commonly used publically available health datasets.}
    \label{tab:datasets}
    \vspace{-0.1in}
\end{table*}

\subsection{Interpretable Predictive Modeling} 

Interpretability is an important property of health risk prediction models because it is related to the life and death of patients when they are applied in real life. Interpretability helps doctors and patients understand the reasoning behind the predictive model and trust its risk prediction or treatment recommendation. Existing interpretability mechanisms designed in the health risk prediction models can be divided into three categories.

\paragraph{Attention-Based Interpretation.} 
The first type of interpretability method utilizes attention weights to explain the importance of different diagnosis codes or visits. This method has been widely used for RNN-based methods~\cite{choi2016retain,ma2017dipole,bai2018interpretable,ma2018health} as we have mentioned above. Attention mechanisms can also be incorporated into Transformer-based predictive models such as HiTANet~\cite{luo2020hitanet} and LSAN~\cite{ye2020lsan}. Take LSAN as an example for illustration. It introduces a hierarchical attention mechanism that assigns flexible attention weights to different diagnosis codes by their relevance to corresponding diseases in the diagnosis code level and pays greater attention to visits with higher relevance in the visit level. Such design provides a fine-grained interpretability. The attention weights in the hierarchy of diagnosis codes and visits can be interpreted into which symptoms and visits are paid more attention to for health risk prediction.

\paragraph{Personalized Knowledge Graph-Based Interpretation.} 
A weakness of attention-based interpretation is that they cannot explicitly express the reasoning path of the health risk prediction models. To tackle that, later research proposes the idea of using a medical knowledge graph to provide explicit reasoning for interpretation. A representative method for interpretability in this type is named MedPath~\cite{ye2021medpath}. MedPath uses  SemMed, a large-scale online medical knowledge graph, to extract personalized knowledge graphs (PKGs) containing all possible disease progression paths from observed symptoms to target diseases. The extracted PKGs can show how the existing observed symptoms from a patient can gradually lead to the target disease, which is more explicit compared to attention weights. Other health risk prediction models using knowledge graphs include the ones designed by~\cite{yang2023kerprint,lyu2023causal,xu2022time}. These methods have shown that knowledge graphs are good candidates for improving performance and interpretability.

\paragraph{Medical Text-Based Explicit Interpretation.}
Another way to provide explicit interpretability is to use unstructured medical text, first proposed in MedRetriever~\cite{ye2021medretriever}. MedRetriever creates a pool of candidate medical text segments from online medical text sources and uses the EHR and target disease document embeddings to dynamically retrieve the related ones that can explain the disease progression from the symptoms to the target disease. By reading the retrieved medical text in the end, both patients and doctors can obtain an easy-to-understand interpretation.

\paragraph{Uncertainty-Based Interpretation.}
Although the aforementioned interpretation methods can provide a decent explanation about how the prediction models work, they still cannot simply say ``I do not know," when they are uncertain in their predictions. However, it is particularly important in clinically relevant tasks. Knowing how confident the prediction is, the clinical experts can trust the results and interpretations with high confidence and set the ones with low confidence aside for another manual inspection. To address this issue, \cite{huang2022trustsleepnet} proposes to use evidential deep learning to quantify the uncertainty. Besides predicting the target class, it also predicts the density of the probability of each class,  which can derive associated uncertainty to express how confident the prediction is.


\section{Benchmarks and Toolkits}\label{sec:4} 

\paragraph{Publicly Available Healthcare Databases.}
We present a summary of commonly used publicly available health datasets in Table~\ref{tab:datasets}. In addition to these widely utilized datasets, certain country-specific datasets play a crucial role in predictive modeling within the healthcare domain. For instance, the Clinical Practice Research Datalink (CPRD, \url{https://cprd.com/}) dataset is a comprehensive, anonymized database of primary care records from the UK. The Health Insurance Review \& Assessment (HIRA) Database (\url{https://www.hira.or.kr/eng/main.do}) is a comprehensive medical database in South Korea, which holds extensive healthcare information for the entire South Korean population. 

\paragraph{Private Healthcare Databases.}
Most extensive healthcare databases operate on a private basis, accessible only to select researchers who can extract pertinent data, as outlined in Table~\ref{tab:sum}. Notably, TriNetX (\url{https://trinetx.com/}) stands out as a notable example, representing a comprehensive dataset that seamlessly integrates real-time access to longitudinal clinical data with cutting-edge analytics. This integration serves to optimize various aspects, including protocol design, feasibility assessments, site selection, patient recruitment, and the generation of real-world evidence. It is noteworthy that the Diamond Network within TriNetX encompasses data from 92 global institutions, capturing information from an expansive cohort of 213,167,071 patients.
Similarly, the IBM MarketScan Research Database mirrors TriNetX in functionality, offering de-identified, longitudinal patient-level claims, and specialty data. This resource boasts coverage for over 273 million unique patients, providing a rich source for diverse healthcare analyses.

\paragraph{Toolkits.}
PyHealth (\url{https://pyhealth.readthedocs.io/en/latest/}) is a comprehensive deep-learning toolkit for predictive modeling, which integrates diverse EHR datasets, such as MIMC,  eICU, and all OMOP-CDM-based databases, and several deep learning algorithms for multiple health-related tasks, e.g., patient hospitalization prediction, mortality prediction, and ICU length stay forecasting. This toolkit allows the users to customize their own pipelines by following the 5 stages: load dataset, define task function, build deep learning models, model training, and inference.
There are also several other toolkits or open-source codes for specific uses such as MedCAT (\url{https://medcat.readthedocs.io/en/latest/}), Fasten (\url{https://github.com/fastenhealth}), MONAI (\url{https://monai.io/}) and NiftyNet (\url{https://niftynet.io/}). 

\section{Open Challenges and Future Directions}\label{sec:5} 

In this section, we delve into various open challenges and propose future research directions across key aspects, including model trustworthiness, data characteristics, model training, collaborative learning paradigms.

\subsection{Trustworthy Predictive Modeling}
Contemporary predictive modeling techniques predominantly prioritize enhancing model performance. While performance is undoubtedly pivotal, equal emphasis must be placed on trustworthiness, encompassing accuracy, reliability, ethics, and transparency. Trustworthy predictive models play a crucial role in supporting healthcare professionals in making informed decisions, improving patient outcomes, and optimizing healthcare processes. Despite its paramount importance, the exploration of trustworthy predictive models in healthcare remains an underexplored domain.

In light of this, we outline the following research directions: 
(1) \emph{LLM-driven interpretable model design}. Our prior work~\cite{promptehr} demonstrates that LLMs encapsulate medical knowledge and possess the potential to interpret model outputs in human-understandable lay language, thereby enhancing model reliability.
(2) \emph{Ethical model design}. Given the diversity within EHR databases, designing fair and robust models while ensuring patient privacy is imperative. A promising research direction involves the development of unbiased and robust predictive models.
(3) \emph{Human-in-the-loop learning}. Existing models primarily rely on data, often neglecting the vital input of healthcare professionals. A human-in-the-loop approach advocates for the inclusion of domain expertise, enabling clinicians to provide feedback and contributing to the trustworthiness of the model.

These research directions collectively pave the way for the development of predictive models that not only excel in performance but also uphold the principles of interpretability, ethics, and collaboration with healthcare professionals, fostering trust and reliability in healthcare applications.

\subsection{Data Scarcity/Sparsity}
Training deep learning models in the healthcare domain often demands a substantial volume of data. However, as indicated in Table~\ref{tab:datasets}, publicly available healthcare data is often limited, especially for rare diseases or specific patient populations. Compounding this issue, EHR data frequently features missing values, leading to substantial data loss during preprocessing. Furthermore, inherent quality challenges in EHR data, such as inaccuracies, duplications, or inconsistencies, further exacerbate the difficulties associated with working with sparse and scarce data. Consequently, acquiring an ample dataset for model training becomes a formidable challenge.
To address this fundamental challenge, one potential solution involves the generation of synthetic EHR data using innovative techniques. Despite recent proposals of models utilizing VAE~\cite{twin,eva}, GAN~\cite{medgan,synteg,maskehr}, diffusion-based models~\cite{meddiff}, and LLM~\cite{promptehr}, these approaches still grapple with the limitation of producing realistic EHR data, specifically in simulating the unique characteristics inherent in EHR datasets.

\subsection{Pretraining across Multiple Data Sources}
Representation learning is a fundamental task in healthcare, particularly in scenarios where labeled data is scarce. Unsupervised or self-supervised learning techniques become essential for accurately capturing health features. While models like GRAM~\cite{choi2017gram} and KAME~\cite{ma2018kame} have been proposed to enhance code representations, they often overlook the multimodal and hierarchical characteristics inherent in EHR data.
In our ongoing work, we are engaged in training a multimodal model named MedHMP~\cite{wang2023hierarchical} on the MIMIC-III dataset. However, this model currently covers only a subset of modalities, leaving others unexplored. To address this limitation and enhance representation learning, a promising avenue for future research involves collaborative pre-training with multiple healthcare datasets from distinct sources. By leveraging shared modalities across these datasets, we can amass sufficient training data to establish a health-specific pre-trained model. This approach holds the potential to comprehensively address the challenge of representation learning across diverse healthcare data sources, accounting for their multimodal and 

\subsection{Federated Training for Foundation Models}
While the transfer of knowledge from a large pre-trained foundation model significantly enhances performance, this approach remains largely unexplored in the context of training foundation models for healthcare. The challenges arise due to the sensitive and private nature of healthcare data, making it infeasible to centrally train such a model.
To address these challenges, a potential solution involves the utilization of advanced federated learning techniques. This approach allows for the collaborative training of a foundation model without necessitating the sharing of stakeholders' private data. Federated learning thus presents a promising avenue for efficiently training healthcare-specific foundation models while respecting privacy constraints.

Nevertheless, the direct application of federated learning in healthcare encounters challenges, primarily stemming from the significant heterogeneity in the distribution of healthcare data, as illustrated in our prior work~\cite{wang2022towards}. Addressing the substantial imbalance and heterogeneity in EHR data across clients represents an open challenge, thereby establishing a promising research direction for the application of federated learning in healthcare.

\section*{Acknowledgements}
This work is partially supported by the National Science Foundation under Grant No. 2238275 and the National Institutes of Health under Grant No. R01AG077016.

\bibliographystyle{named}
\bibliography{ijcai24}

\end{document}